# Terrain-Perception-Free Quadrupedal Spinning Locomotion on Versatile Terrains: Modeling, Analysis, and Experimental Validation

*Hongwu Zhu[1,2], Dong Wang[1,2], Nathan Boyd[3], Ziyi Zhou[3], Lecheng Ruan[4], Aidong Zhang[1,2], Ning Ding[1,2], Ye Zhao[3]\* and Jianwen Luo[1,2]\**

[1]*Shenzhen Institute of Artificial Intelligence and Robotics for Society (AIRS), Shenzhen, China,* [2]*Institute of Robotics and Intelligent Manufacturing (IRIM), The Chinese University of Hong Kong (CUHK), Shenzhen, China,* [3]*George W. Woodruff School of Mechanical Engineering, Georgia Institute of Technology, Atlanta, GA, United States,* [4]*Beijing Institute for General Artificial Intelligence, Beijing, China*

Dynamic quadrupedal locomotion over rough terrains reveals remarkable progress over the last few decades. Small-scale quadruped robots are adequately flexible and adaptable to traverse uneven terrains along the sagittal direction, such as slopes and stairs. To accomplish autonomous locomotion navigation in complex environments, spinning is a fundamental yet indispensable functionality for legged robots. However, spinning behaviors of quadruped robots on uneven terrain often exhibit position drifts. Motivated by this problem, this study presents an algorithmic method to enable accurate spinning motions over uneven terrain and constrain the spinning radius of the center of mass (CoM) to be bounded within a small range to minimize the drift risks. A modified spherical foot kinematics representation is proposed to improve the foot kinematic model and rolling dynamics of the quadruped during locomotion. A CoM planner is proposed to generate a stable spinning motion based on projected stability margins. Accurate motion tracking is accomplished with linear quadratic regulator (LQR) to bind the position drift during the spinning movement. Experiments are conducted on a small-scale quadruped robot and the effectiveness of the proposed method is verified on versatile terrains including flat ground, stairs, and slopes.

Keywords: quadruped robot, turning gait, spinning locomotion, trajectory tracking control, versatile terrains

## 1 INTRODUCTION

Quadruped robots, equipped with advanced walking ability over unstructured terrains, have started to make their way into human environments (Ijspeert, 2014; Yang et al., 2020; Bledt and Kim, 2020). The current quadruped robots can mimic not only static gaits of animals but also highly agile and dynamic behaviors, such as galloping, jumping, and back-flipping (Katz et al., 2019; Kim et al., 2019), which enable them to traverse unstructured terrains (Bledt et al., 2018; Kim et al., 2020; Jenelten et al., 2020). Yet, certain locomotion behaviors have not been explored, e.g., the circular spinning locomotion (Carpentier and Wieber, 2021). Dogs often spin to inspect the environment and search for potential threats (Park et al., 2005; Chen et al., 2017). For the robot counterpart, spinning gait is also an indispensable component to fulfill for trajectory tracking tasks in autonomous navigation (Xiao et al., 2021), because any curves can be decoupled into forward, lateral, and spinning locomotions (Ma et al., 2005; Wang et al., 2011; Hong et al., 2016). However, the highly dynamic spinning is still challenging due to the complex dynamics, such as uncertain contact, inaccurate foot placement, and potential tripping (Ishihara et al., 2019). Consequently, it is significant to investigate a method that can accomplish the accurate spinning locomotion over complex terrains.

Currently, most legged robots generate spinning motions by manipulating with yaw joints on pelvis or waist. (Miao and Howard, 2000) proposed a tripod turning gait for a six-legged walking robot by tuning the appropriate motion trajectory of the supporting leg relative to the robot body in simulation. (Roy and Pratihar, 2012) focused on improving turning gait parameters to minimize the energy consumption of a six-legged walking robot. Estremera et al. (2010) analyzed and formulated a spinning crab gait for a six-legged walking robot over rough terrain. Park et al. (2005) proposed a spinning gait for a quadruped walking robot with a waist joint, but the robot could not walk with the spinning gait on rough terrain. Chen et al. (2017) introduced a tripod gait-based turning gait of a six-legged walking robot. (Mao et al., 2020) demonstrated the Hexa-XIII robot with 12-leg joint motors and 1 waist motor. The six-legged robot improves the stability and decreases the leg interference for spinning compared with the common tripod gaits. However, the aforementioned turning/spinning gaits that are based on stability margin all belong to the static gait planning, which is only available for low-speed walking (Hong et al., 2016).

In the meantime, quadrupedal hardware has advanced significantly to enable highly mobile and agile motions. For example, the MIT Cheetah achieved a high speed of 3.7 m/s for straight running (Kim et al., 2019). The MIT mini Cheetah robot is capable of accomplishing highly dynamic motions, including trotting, running, bounding, and back-flipping (Bledt et al., 2018; Kim et al., 2019). These quadruped robots have 3 degrees of freedom (DoF) on each leg, but without rotational DoF in the pelvis (Estremera and Gonzalez, 2002; Ma et al., 2005). This leg configuration becomes mainstream on current quadruped robots due to better bionics in geometric topology. In this case, the spinning locomotion can be only realized through the rolling of the spherical foot-ends on the ground (Miura et al., 2013; Yeon and Park, 2014), which leads to the gait instability and CoM drift.

To address this challenge, this study first proposes a gait planning method with a modeled spherical foot for turning and spinning in the trotting gait. Based on the geometrical relationship of the foot end effector and body coordinate, a desired turning foot position is generated (Palmer and Orin, 2006; Roy and Pratihar, 2012; Liu et al., 2017). A spinning gait is obtained when the turning radius becomes zero. The CoM trajectory is generated directly by mapping from the planned foot positions. Second, a linear quadratic regulator (LQR) feedback controller is devised to compensate the cumulative errors along the trajectory to track the fixed point under a small turning radius (Thrun et al., 2009; Xin et al., 2021). The proposed method is validated on a quadruped robot platform for spinning over versatile terrains, and the results show improved convergence and stability when spinning with a trotting gait on challenging terrains. The main contributions of this letter lie in the following threefold:

i) Devise a turning/spinning gait planner with foot end effector kinematic correction and a CoM trajectory planner based on generalized support polygon.
ii) Devise a LQR controller to guarantee the spinning radius to be strictly bounded.
iii) Conduct experimental validations of the quadruped robot with satisfactory locomotion performance.

The rest of this article is organized as follows. **Section 2** introduces the overall framework of this study. A turning/spinning step planner with a foot-end effector kinematic correction. A legged odometry feedback planner based on the LQR technique is introduced in **Section 3** to guarantee the spinning movement to be bounded within a limited range. Simulation and experiment results are shown in **Section 4**. **Section 5** concludes this study.

## 2 FRAMEWORK

In order to achieve terrain-perception-free yet accurate spinning locomotion on versatile terrains, this study proposes a control framework as shown in **Figure 1**. This control framework incorporates the MIT mini cheetah controller as the low-level motion control module (Kim et al., 2019), which consists of the model predictive control (MPC) and whole-body control (WBC) modules. The robot's state estimator and kinematics/dynamics model is used to obtain the current position, velocity, acceleration of the CoM and joints, respectively, using a linear Kalman filter. The errors of the foot rolling are taken into account in the motion planning process, and the kinematics of the legs is corrected by the foot end effector kinematic modification method (FKM). The proposed LQR controller is used to generate the body control commands, where the tracking error of the trajectory is strictly bounded. With the leg kinematics correction, the resultant body position and velocity are sent to MPC and WBC to calculate the expected position, velocity, and torques for joint actuators (Luo et al., 2019). The MPC computes the optimal reaction forces over a time horizon with a linearized single rigid body template model. The WBC tracks the computed reaction forces generated from the MPC for uncontrollable maneuvers such as galloping. These modules including MIT controller, projected support polygon (PSP), CoM trajectory planner, FKM, and LQR form our accurate spinning control framework (ASC).

Since the foot-end effector of the robot is spherical, the foot-end effector rolls on the ground as the leg posture changes. For small-scale quadruped robots, the ratio between radius of ball foot and shank length is large. As a result, the large radius foot will change the contact point and CoM position as the robot spins around the yaw axis during the support phase as shown in **Figure 2**. This deviation is not negligible during a highly agile locomotion and the spherical contact engagement needs to be investigated and modeled.

Additionally, in order to further guarantee the accuracy of the locomotion, a method of planning the trajectory of the CoM that mitigates translational drifting is developed. During the double support of the robot, the CoM drift is difficult to avoid. Once the CoM shifts from the diagonal of the support foot point, additional torque is applied by the gravity and affects the stability of the robot. On unstructured terrains, there are frequent undesired ground contacts due to the unpredictability and complexity. To improve the performance, the slope of the terrain is estimated based on the location of the feet. By mapping from the next foothold, the CoM position is adjusted to ensure motion feasibility based on PSP.

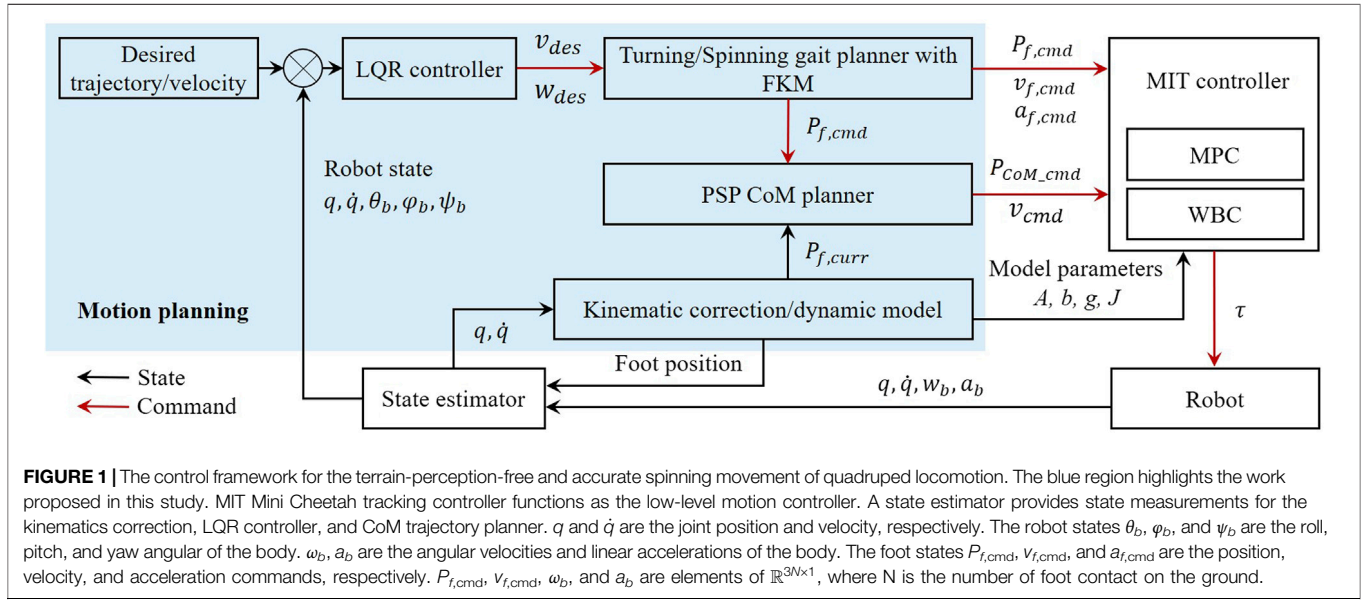

FIGURE 1 | The control framework for the terrain-perception-free and accurate spinning movement of quadruped locomotion. The blue region highlights the work proposed in this study. MIT Mini Cheetah tracking controller functions as the low-level motion controller. A state estimator provides state measurements for the kinematics correction, LQR controller, and CoM trajectory planner. $q$ and $\dot{q}$ are the joint position and velocity, respectively. The robot states $\theta_b$, $\varphi_b$, and $\psi_b$ are the roll, pitch, and yaw angular of the body. $\omega_b$, $a_b$ are the angular velocities and linear accelerations of the body. The foot states $P_{f,cmd}$, $v_{f,cmd}$, and $a_{f,cmd}$ are the position, velocity, and acceleration commands, respectively. $P_{f,cmd}$, $v_{f,cmd}$, $\omega_b$, and $a_b$ are elements of $\mathbb{R}^{3N \times 1}$, where N is the number of foot contact on the ground.

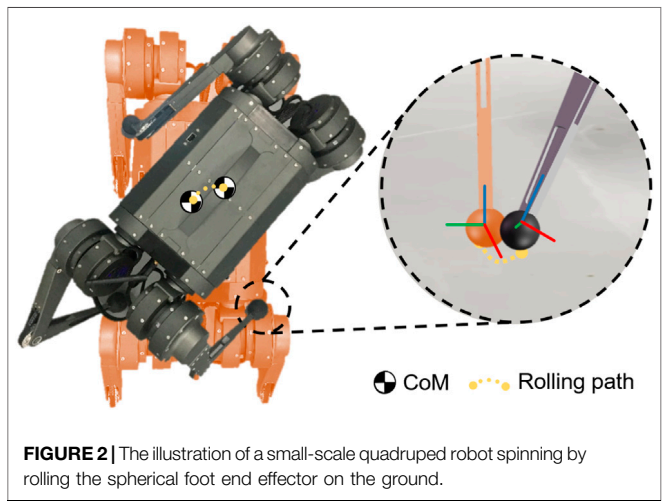

FIGURE 2 | The illustration of a small-scale quadruped robot spinning by rolling the spherical foot end effector on the ground.

# 3 GAIT AND CENTER OF MASS TRAJECTORY PLANNING FOR SPINNING LOCOMOTION

In this section, a turning/spinning gait planner with foot-end effector kinematic modification (FKM), a CoM planner based on projected support polygon (PSP), and a CoM trajectory tracker based on an LQR controller are introduced respectively.

## 3.1 Turning/Spinning Gait Planner and Foot End Effector Kinematic Modification

As shown in **Figure 3**, the angle $\gamma$ represents the circle angle in the turning process from the point A to the point B. Therefore, the translation variation of the support leg relative to the body of the robot between A and B is the variation of the CoG of the robot relative to the forward direction of the $x$ axis and the lateral direction of the $y$ axis.

Let $\Delta l_{x,t}$ and $\Delta l_{y,t}$ be the variation which is given as follows:

$$\Delta \mathbf{l}_t = \begin{bmatrix} \Delta l_{x,t} \\ \Delta l_{y,t} \\ \Delta l_{z,t} \end{bmatrix} = \begin{bmatrix} R \sin \gamma \\ R(1 - \cos \gamma) \\ 0 \end{bmatrix}. \quad (1)$$

The hip position of right front (RF) leg in the body of the robot coordinate system is $(L/2, -W/2)$, where $L$ and $W$ are the length and width of the robot body, respectively, because the body rotates $\gamma$ angle in the counterclockwise direction. In the moment, the support legs are all right below the hip as shown in **Figure 3A**. The rotation variation of the hip of the body is also the variation of the support leg in the plane coordinate system. Therefore, the variation of the hip of the robot relative to the body rotation ($\Delta l_r$) can be obtained as follows:

$$\Delta \mathbf{l}_r = \begin{bmatrix} \Delta l_{x,r} \\ \Delta l_{y,r} \\ \Delta l_{z,r} \end{bmatrix} = \begin{bmatrix} \frac{L}{2} \cos \gamma + \frac{W}{2} \sin \gamma \\ \frac{L}{2} \sin \gamma - \frac{W}{2} \cos \gamma \\ 0 \end{bmatrix}. \quad (2)$$

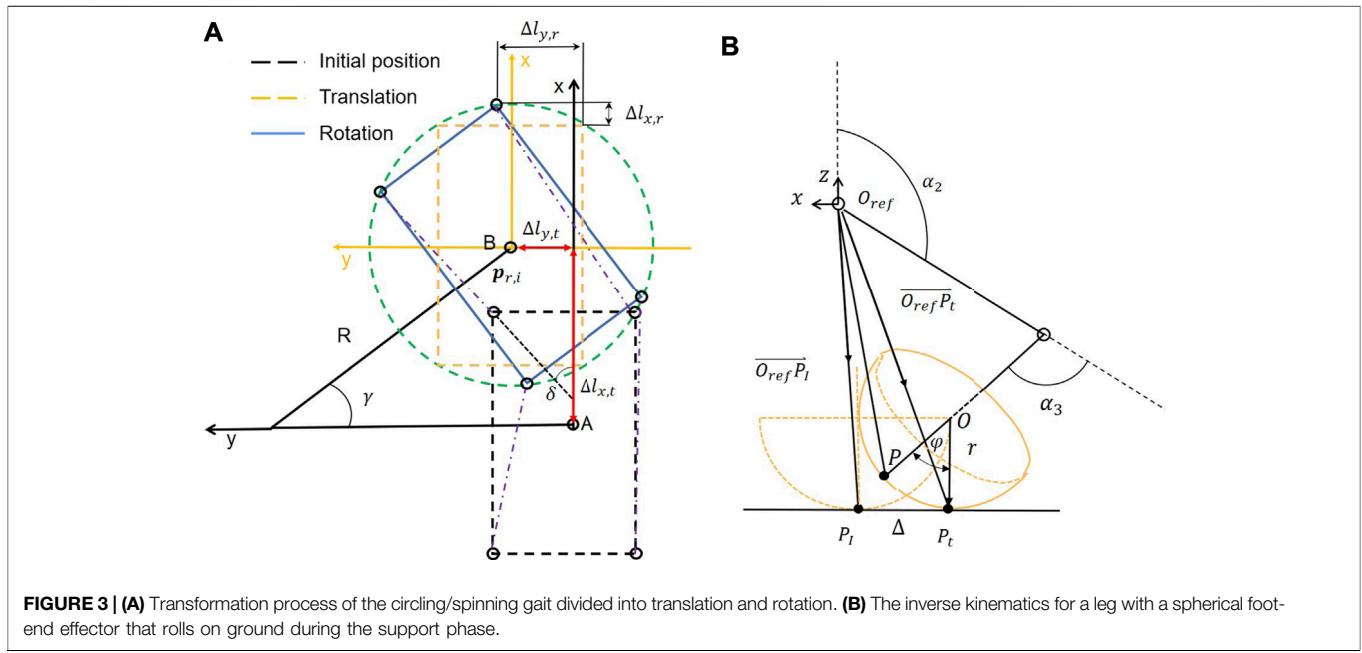

FIGURE 3 | (A) Transformation process of the circling/spinning gait divided into translation and rotation. (B) The inverse kinematics for a leg with a spherical foot-end effector that rolls on ground during the support phase.

Based on the translation variation and rotation variation equations, the expression of the moving foot step of support legs with respect to the body coordinate system in the initial state can be obtained as follows:

$$\Delta \mathbf{l} = \Delta \mathbf{l}_t + \Delta \mathbf{l}_r = \begin{bmatrix} R\sin\gamma + \frac{L}{2}\cos\gamma + \frac{W}{2}\sin\gamma \\ R(1-\cos\gamma) + \frac{L}{2}\sin\gamma - \frac{W}{2}\cos\gamma \\ 0 \end{bmatrix}. \quad (3)$$

The sum of the current projection position of the hip joint and the calculated step length is used to plan the next footholds, which is given as follows:

$$\mathbf{P}_{f,cmd} = \mathbf{P}_{shoulder,i} + \Delta \mathbf{l}. \quad (4)$$

Due to the relative rolling between the spherical foot end and the ground surface, the contact point will constantly change and the movement trajectory of the body deviates from the desired trajectory. The deviation caused by the spherical end effector occurs not only in the vertical direction but also in the horizontal direction, which consequently leads to a severe tracking error and even locomotion failure. Therefore, it is necessary to propose a kinematics correction algorithm to eliminate this deviation.

Regardless of the shape and volume of the foot, the foot position vector $p$ can be obtained by the forward kinematic as follows:

$$\hat{p} = \begin{bmatrix} s_{23}L_3 + s_2L_2 \\ s_1c_{23}L_3 + s_1c_2L_2 \\ -c_1c_{23}L_3 - c_1c_2L_2 \end{bmatrix}, \quad (5)$$

where $s_i = \sin\alpha_i$, $c_i = \cos\alpha_i$, $s_{ij} = \sin(\alpha_i + \alpha_j)$, and $c_{ij} = \cos(\alpha_i + \alpha_j)$ and $\alpha_i$, and $\alpha_j$ are the $i$th and $j$th joint angles as shown in **Figure 3B**, respectively.

Similarly, the inverse kinematics solution is obtained through the leg kinematics which is denoted as follows:

$$\alpha = \begin{bmatrix} \alpha_1 \\ \alpha_2 \\ \alpha_3 \end{bmatrix} = \begin{bmatrix} \arctan\frac{\hat{P}_y}{\hat{P}_x} \\ \arcsin\frac{A + L_2^2 - L_3^2}{2L_2\sqrt{A}} - \arctan\frac{\sqrt{A - (\hat{P}_x)^2}}{\hat{P}_z} \\ \pm\arccos\frac{A - L_2^2 - L_3^2}{2L_2L_3} \end{bmatrix}, \quad (6)$$

where $A = (\hat{P}_x)^2 + (\hat{P}_y)^2 + (\hat{P}_z)^2$. $\alpha_1$, $\alpha_2$, $\alpha_3$ represents the hip joint angle, thigh joint angle, and calf joint angle, respectively.

Even if no slip occurs, the contact point is constantly changing and the body CoM deviates from the desired trajectory as shown in **Figure 3B** and **Supplementary Video S1**. This deviation is attributed to the ball foot-end effector roll as the body moves during the support phase (Guardabrazo et al., 2006). In order to eliminate this modeling error, the required joint rotation angles need to be corrected to eliminate the mismatch between the point-foot model and ball foot (Kwon and Park, 2014). The ideal point-foot position relative to the hip joint coordinate system is derived by the forward dynamics in the following way (Lavaei et al., 2017):

$$|\Delta| = |\vec{P_tP_I}| = |\widehat{P_tP}|, \quad (7)$$

where $|\widehat{P_tP}|$ is the arc length between the foot reference point $P$ and the real contact point $P_t$. $P$ and $P_I$ are the same point at the initial contact state. Assuming there is no slip, the displacement offset of the foot on the ground is equivalent to the rotated distance on the foot. As shown in **Figure 3B**, the real foothold is obtained as follows:

$$\vec{O_{ref}P_t} = \begin{bmatrix} -L_3 s_{23} - L_2 s_2 \\ -L_3 s_1 c_{23} - L_2 s_1 c_2 \\ L_3 c_1 c_{23} + L_2 c_1 c_2 - r \end{bmatrix}, \quad (8)$$

where $r$ represents the radius of spherical foot-end effector. For the ideal foothold, we have the following:

$$\vec{O_{ref}P_I} = \vec{O_{ref}P_t} + \Delta = \begin{bmatrix} -L_3 s_{23} - L_2 s_2 - \Delta_x \\ -L_3 s_1 c_{23} - L_2 s_1 c_2 - \Delta_y \\ L_3 c_1 c_{23} + L_2 c_1 c_2 - r \end{bmatrix}, \quad (9)$$

where $\Delta_x$, $\Delta_y$ represents the vector $\Delta$ in the $x$ and $y$ directions of the base reference coordinate system. Therefore, the angle $\phi$ between the third linkage and the perpendicular of the horizontal plane can be obtained and $\Delta_z = 0$, $\Delta$, and $\vec{O_{ref}P}$ are coplanar; therefore, we have the following:

$$\Delta = \begin{bmatrix} \Delta_x \\ \Delta_y \\ 0 \end{bmatrix} = \begin{bmatrix} \dfrac{-r s_{23} \varphi}{\sqrt{s_1^2 c_{23}^2 + s_{23}^2}} \\ \dfrac{-r s_1 c_{23} \varphi}{\sqrt{s_1^2 c_{23}^2 + s_{23}^2}} \\ 0 \end{bmatrix}, \quad (10)$$

where $\varphi = \arccos(-c_1 c_{23})$ and $|\Delta| = r\varphi$.

Hence, the kinematic solution to the ideal foothold in the base–joint coordinate system can be obtained as follows:

$$\vec{O_{ref}P_I} = \begin{bmatrix} P_{Ix} \\ P_{Iy} \\ P_{Iz} \end{bmatrix} = \begin{bmatrix} -L_2 s_{23} - L_3 s_2 - \dfrac{-r s_{23} \varphi}{\sqrt{s_1^2 c_{23}^2 + s_{23}^2}} \\ -L_3 s_1 c_{23} - L_2 s_1 c_2 - \dfrac{-r s_1 c_{23} \varphi}{\sqrt{s_1^2 c_{23}^2 + s_{23}^2}} \\ L_3 c_1 c_{23} + L_2 c_1 c_2 - r \end{bmatrix}. \quad (11)$$

For the single leg with a spherical foot end, the position of the ideal foothold point in the root joint coordinate system is known. The rotation angle vector of each joint of the leg can also be solved through the following inverse kinematics:

$$\alpha' = \begin{bmatrix} \alpha_1' \\ \alpha_2' \\ \alpha_3' \end{bmatrix} = \begin{bmatrix} \arctan \dfrac{P_{Iy} - \Delta_y}{P_{Iz} + r} \\ \arcsin \dfrac{A' + L_2^2 - L_3^2}{2 L_2 \sqrt{A'}} - \arctan \dfrac{\sqrt{A' - (P_{Ix} + \Delta_x)^2}}{P_{Ix} + \Delta_x} \\ \pm \arccos \dfrac{A' - L_2^2 - L_3^2}{2 L_2 L_3} \end{bmatrix}, \quad (12)$$

where $A = (\hat{P}_{Ix} + \Delta_y)^2 + (\hat{P}_{Iy} + \Delta_y)^2 + (\hat{P}_{Iz} + r)^2$. $\alpha_1'$, $\alpha_2'$, $\alpha_3'$ represents the hip joint angle, thigh joint angle, and calf joint angle, respectively.

Besides, a terrain estimation method is devised for uneven terrains by taking the height difference of the four legs into account. The terrain height can be modeled using the following linear regression:

$$z(x, y) = a_0 + a_1 x + a_2 y. \quad (13)$$

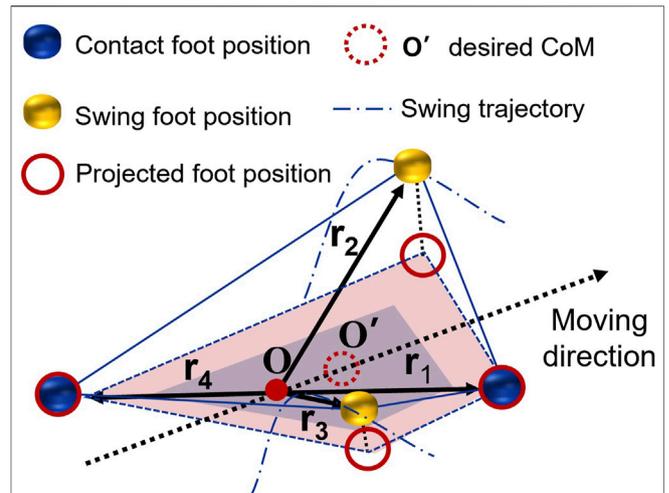

**FIGURE 4** | The illustration of the desired CoM trajectory calculation is based on the PSP CoM planner.

Coefficients $\mathbf{a} = (a_0, a_1, a_2)^T$ of (**Eq. 13**) are obtained through the solution of the minimum squares problem as is described in a study by ((Bledt et al. (2018)) which are given as follows:

$$\mathbf{a} = (\mathbf{W}^T \mathbf{W})^{-1} \mathbf{W}^T \mathbf{p}_c^z, \quad (14)$$

where $\mathbf{p}_c = (\mathbf{p}_c^x, \mathbf{p}_c^y, \mathbf{p}_c^z)^T$ is the most recent contact point of each foot, and $\mathbf{W} = \begin{bmatrix} 1 & \mathbf{p}_c^x & \mathbf{p}_c^y \end{bmatrix}_{4 \times 3}$. When the robot encounters uniformly changing terrains such as block roadblocks and stairs, this modeling method is still effective. In this way, the terrain information has been roughly estimated to assist in the modification of the upcoming footstep location. The body posture angle of the robot will be adjusted according to the angle of the ground plane in (**Eq. 13**) to adapt to the terrain.

When the robot walks on unstructured terrain, the estimated terrain is combined to modify the current planned position. The upcoming footstep location is shown as follows:

$$\mathbf{P}_{f,cmd} = \begin{bmatrix} 1 & 0 & 0 \\ 0 & 1 & 0 \\ a_1 & a_2 & 1 \end{bmatrix} \mathbf{P}_{f,cmd} + \begin{bmatrix} 0 \\ 0 \\ a_0 \end{bmatrix}, \quad (15)$$

where $a_0$, $a_1$, $a_2$ are obtained through the solution of the least squares problem as mentioned above. When the robot is walking on a plane, using (**Eq. 15**) to calculate the next footing point is an effective method. However, when the robot is traversing on unstructured terrain, the upcoming footstep location needs to be modified so that the actual foot-end effector trajectory of the quadruped robot can track the planned trajectory.

## 3.2 Center of Mass Planner Based on Projected Support Polygon

A majority of studies in turning gaits belong to the static gait planning with a slow walking speed because the gaits are optimized based on stability margin (SM) to ensure the balance (Chen et al., 2017; Luo et al., 2021). SM is the shortest distance from the vertical projection of the CoM to any point on the boundary of the support polygon pattern. For dynamics gait like trotting of quadruped robots, the two supporting point feet cannot form conventional polygon patterns (Luo et al., 2020). Here, we calculate the desired CoM trajectory by introducing the PSP concept, mapping the foot position of the swing leg as a virtual vertex (**Figure 4**).

The midpoint of diagonal line of two supporting feet is marked as $O$. Four vectors $\mathbf{r}_i \in \{FR: 1, FL: 2, BR: 3, BL: 4\}$ start from $O$, pointing to the position of each foot point. Then, four virtual vectors can be obtained by projecting on the ground.

Instead of uniform interpolating centroid positions based on the velocity at the current and desired centroid positions, a set of weights are used to calculate foot position in the swing phase. The weights $P$ obey common unimodal distributions like geometric, Poisson, or Gaussian distribution.

$$P(c|s_\phi, \phi) = D(s_\phi, \phi), \tag{16}$$

where $P(c|s_\phi, \phi) \in [0, 1]$ corresponds to the adaptive weighting factor during the scheduled stance and swing phase. The phase $\phi$ represents the gait phase, and $s_\phi$ acts as a switch between swing ($P(c|\phi) = 0$) or stance ($P(c|\phi) = D(s_\phi)$). The closer the leg is to the middle of the stance phase, the heavier the coefficient $P(c|s_\phi, \phi) = D(s_\phi, \phi)$ of the support foothold location. On the contrary, the closer the leg is to the middle of swing phase, the smaller the $P(c|s_\phi, \phi) = D(s_\phi, \phi)$ of the foothold location is.

$$\mathbf{V}_i = P(i, \phi) \cdot \hat{\mathbf{r}}_i. \tag{17}$$

$\mathbf{V}_i$ is the vertex of the foothold location after multiplying the weights. Four projected supporting vertexes $\mathbf{P}_i$ can be obtained from $\mathbf{V}_i$. Given the average value of the vertices, the expected value of the robot's expected CoM value is approximated as follows:

$$\begin{cases} \hat{\mathbf{p}}_{CoM,i} = \frac{1}{N}\sum_{i=1}^{N}\mathbf{P}_i, \\ \hat{\mathbf{v}}_{CoM} = \dot{\hat{\mathbf{p}}}_{CoM}. \end{cases} \tag{18}$$

The difference between the planned CoM position $\hat{\mathbf{p}}_{CoM,i}$ and the current CoM position $\hat{\mathbf{p}}_{CoM,curr}$ divided by the gait cycle $T$ is the desired velocity. Adding the current CoM by the product of the average velocity $\hat{\mathbf{v}}_{CoM}$ and the unit time $\delta t$ position, we interpolated the CoM trajectory of $f$ points between the current CoM position and the planned CoM position $\hat{\mathbf{p}}_{CoM} = [\hat{\mathbf{p}}_{CoM,1}, \hat{\mathbf{p}}_{CoM,2}, \ldots, \hat{\mathbf{p}}_{CoM,f}]^T$ and sent the continuous CoM position and velocity trajectories (the velocity one is calculated by differentiating the position trajectory) to the MPC and WBC controllers.

## 3.3 Center of Mass Trajectory Tracking

Searching methods are common for path tracking problems of mobile robots. The goal point and path curvature connecting to the goal point are calculated in every step. The goal point $\mathbf{p}_{r,i} = [p_{r,i,x}, p_{r,i,y}]^T$ is illustrated in **Figure 3**. The legs' steering angle $\delta$ can be determined using only the goal point location and the angle between the vehicle's heading vector and the look-ahead vector. The search for goal point $\mathbf{p}_{r,i}$ is determined from the CoM position without look-ahead distance to the desired path ($\mathbf{L}_r$). The distance between the points on the desired path with the current CoM position $\mathbf{p}$ is calculated by the Euclidean distance. The index $i$ and nearest point on the path $\mathbf{p}_{r,i}$ can be obtained. $\theta_r$ is the reference yaw angle of body in the world coordinate. The angular velocity of body is $\omega$. The steering angle $\delta$, the angle between the leg trajectory, and the $x$ axis of body can be determined by the tangent angle of the goal point. The curvature of a circular arc of goal point can be calculated directly.

$$\begin{cases} \mathbf{p}_{r,i} = \arg\min_i \|\mathbf{L}_r - \mathbf{p}\|_2, \\ \theta_r = \arctan(\dot{\mathbf{p}}_{r,i}), \\ R = \frac{(1+\dot{\mathbf{p}}^2)^{(3/2)}}{\ddot{y}}. \end{cases} \tag{19}$$

The generalized ball foot error obtained in the previous section is regulated with a LQR controller. $\mathbf{p}$ is the CoM position and $\gamma$ is the attitude angle of the body. Define state vector $\mathbf{X} = [\mathbf{p}^T, \gamma]^T$ and control vector $\mathbf{u} = [\mathbf{v}^T, \dot{\delta}]$, the body dynamics are formulated as follows:

$$\begin{cases} \dot{x} = v\cos\gamma, \\ \dot{y} = v\sin\gamma, \\ \dot{\gamma} = \omega. \end{cases} \tag{20}$$

By defining $\tilde{\mathbf{X}} = \mathbf{X} - \mathbf{X}_r$, $\tilde{\mathbf{u}} = \mathbf{u} - \mathbf{u}_r$, and linearizing the dynamics around the reference point, the system governing equation is reformulated as follows:

$$\dot{\tilde{\mathbf{X}}} = \mathbf{A}\tilde{\mathbf{X}} + \mathbf{B}\tilde{\mathbf{u}}, \tag{21}$$

where $A$ and $B$ are given as follows:

$$\mathbf{A} = \begin{bmatrix} 0 & 0 & -v_r\sin\gamma \\ 0 & 0 & v_r\cos\gamma \\ 0 & 0 & 0 \end{bmatrix}, \quad \mathbf{B} = \begin{bmatrix} \cos\gamma & 0 \\ \sin\gamma & 0 \\ \tan\gamma & \frac{v_r}{\cos\delta} \end{bmatrix}, \tag{22}$$

where $v_r$ is the desired velocity on $\mathbf{p}_{r,i}$. For controller implementation, (**Eq. 21**) is discretized with the forward Euler discretization:

$$\dot{\tilde{\mathbf{X}}}(k) = \frac{\tilde{\mathbf{X}}(k+1) - \tilde{\mathbf{X}}(k)}{\Delta t}. \tag{23}$$

Then the LQR controller is obtained by minimizing the performance index:

$$\mathbf{J} = \sum_{k=1}^{\infty} (\tilde{\mathbf{X}}^T(k)\mathbf{Q}\tilde{\mathbf{X}}(k) + \tilde{\mathbf{u}}^T(k)\mathbf{R}\tilde{\mathbf{u}}(k)), \tag{24}$$

where positive definite matrices $Q$ and $R$ are weighting parameters.

## 4 SIMULATION AND EXPERIMENT RESULTS

To validate the proposed method, three sets of experiments are conducted in simulations and experiments: the feasible spinning locomotion of trotting gait, the bounded small radius of spinning, and spinning on the slopes and stairs. While our ASC method is generalizable to model any turning action, we primarily focus on showing its effectiveness on fast spin maneuvers over various terrains, where the motion is prone to failures. The experiments are tested on a real small-scale quadruped robot platform.

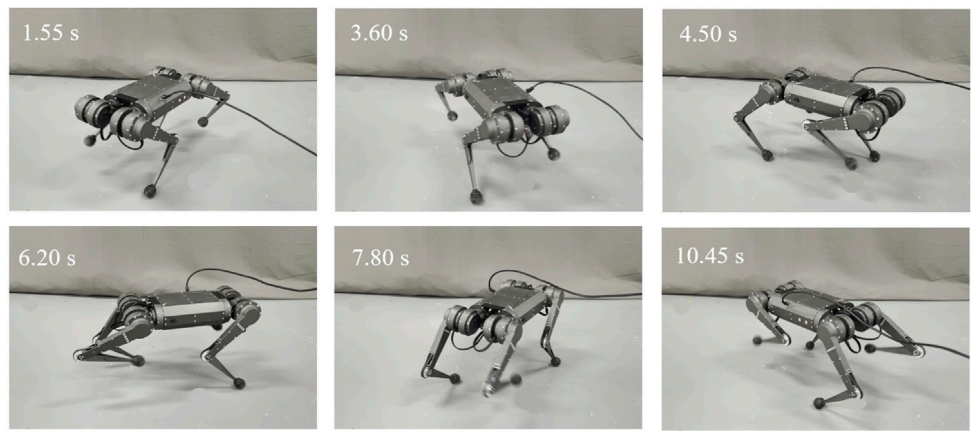

FIGURE 5 | Screenshots of the quadruped robot spinning on the flat ground with an ACS controller.

## 4.1 Experiment Platform

The experiment platform for the spinning test is a small-scale quadruped robot, which is electrically actuated with 12 degrees of freedom, 9 kg weight, and 28 cm tall. The body clearance is 29 cm and length is 38 cm, and the length of thigh and calf joint is 21.5 and 20 cm, respectively. The radius $r$ of foot is 2.25 cm. The locomotion controller is executed on an Intel UP board low-power single-board computer, with a quad-core Intel Atom CPU, 4 GB RAM. Linux with the CONFIG PREEMPT RT patch works as the operating system. UP board is used to run the low-level controller, including MPC, WBC, and the state estimator.

## 4.2 Experimental Validation of Spinning on the Flat Ground

The above method is validated through comparative experiments. The robot is expected to spin at trotting gait on the flat ground. The velocity of the robot in the $x$ and $y$ directions is 0 m s$^{-1}$. The angular velocity $\omega$ is 0.7 rad s$^{-1}$. The gait planner, FKM, PSP CoM planner, and LQR controller are verified for spinning both on simulation and the quadruped platform. The experiment screenshots of the quadruped robot spinning on the flat ground are shown in **Figure 5**.

The CoM trajectories during spinning are shown in **Figure 6**. The PSP CoM planner was used by default in each trial to avoid falling. Eight cycles' data containing about 100 steps were recorded. The results of first 5 seconds were removed, when the robot went straight to the preset position. **Figure 6A** shows the simulated results of different control methods. The black line is the trajectories of MIT controller with a circle having a radius of 2.79 cm, and the trajectory variance is 0.57 mm$^2$. Based on the MIT controller, FKM method is added, and the corresponding trajectories are brown lines. The brown circle has a radius of 1.4 cm with a variance of 0.43 mm$^2$. In our ASC framework, an LQR controller is also added, together with an MIT controller and FKM, to further reduce the radius and bound the trajectories to the origin point. The red lines are the trajectories formed by using our ASC method. The radius reduces to 1.12 cm and the trajectory variance is 0.31 mm$^2$, which clearly shows an improvement in tracking accuracy. **Figure 6B** shows the experimental results on the Mini Cheetah quadruped hardware platform. Though the CoM trajectories have a clear stochastic disturbance compared to simulation, the results show similar features. By using ACS, the CoM trajectory of the robot that spins converges to the fixed point with a radius of 3.84 cm (variance: 0.56 cm$^2$). After adding FKM, the CoM trajectory reaches an intermediate level with a radius of 4.28 cm (variance: 0.5 cm$^2$). With merely an MIT controller, the radius of the CoM trajectory increase to 7.67 cm (variance: 2.50 cm$^2$) and shows an inconsistent tracking performance. In addition, spinning is conducted by using merely LQR and MIT controller in **Supplementary Figure S5**. LQR tends to bind the radius to zero directly, and the trajectory crosses the origin repeatedly. Based on the four sets of comparative experiment, we consider that the components in our ASC framework have different functionalities: i) PSP CoM planner component projects the CoM onto the diagonal of the supporting foot to avoid falling during spinning, which is used by default in our spinning results. ii) FKM eliminates the position error by modeling the mismatch of the point-foot assumption and the ball foot in practice. iii) By incorporating with the LQR, systematic errors are further reduced and a bound is established on the robot's absolute position.

**Figure 7** shows the drift, velocity, and the attitude of the $x$ and $y$ axes during the spinning. 10 s' records containing about 20 steps were recorded. In **Figure 7A**, the $x$ (3.49 cm) and $y$ (1.96 cm) axes drift with an MIT controller is 2 times larger than the drift ($x$: 0.62 cm, $y$: 0.71 cm) using our ASC method in simulation. The drift is also closer to the origin in the world coordinate system. **Figure 7B** represents the drift of the $x$ and $y$ axes on the quadruped hardware platform. Similar to simulation, the fluctuation range of the $x$ (1.25 cm) and $y$ (1.06 cm) axis drifts is small (while the drifts fluctuation range of the *x* and *y* axes is (3.22 cm) and (2.77 cm) and fluctuating around 0, which is beneficial for the center of the robot spinning closer to the origin in the world coordinate system. Besides the effective tracking of the desired CoM point during the robot spinning, the stability of the robot during the spinning is also improved. As shown in **Figure 7**, the roll angle, pitch angle, linear acceleration, and angular acceleration of the robot are recorded. The accuracy of roll and pitch in the dynamic motion is crucial. Large roll and pitch angle variations will cause the robot to tilt or even fall. With our ASC method, the experiment has smaller fluctuations in roll and pitch. The pitch angle of body ranges from -0.02 to 0 rad, and shows a smaller drift from 0 rad in simulation. In the quadruped platform experiment, the calculated mean angle and variance are $1.77 \times 10^{-3}$ rad, $1 \times 10^{-4}$ rad for pitch, and $-1.35 \times 10^{-2}$ rad and $1.17 \times 10^{-4}$ rad for roll, compared with the $-1.75 \times 10^{-3}$ rad, $5.85 \times 10^{-4}$ rad for pitch, and $1.83 \times 10^{-2}$ rad and $3.70 \times 10^{-4}$ rad for roll with using MIT controller methods, respectively.

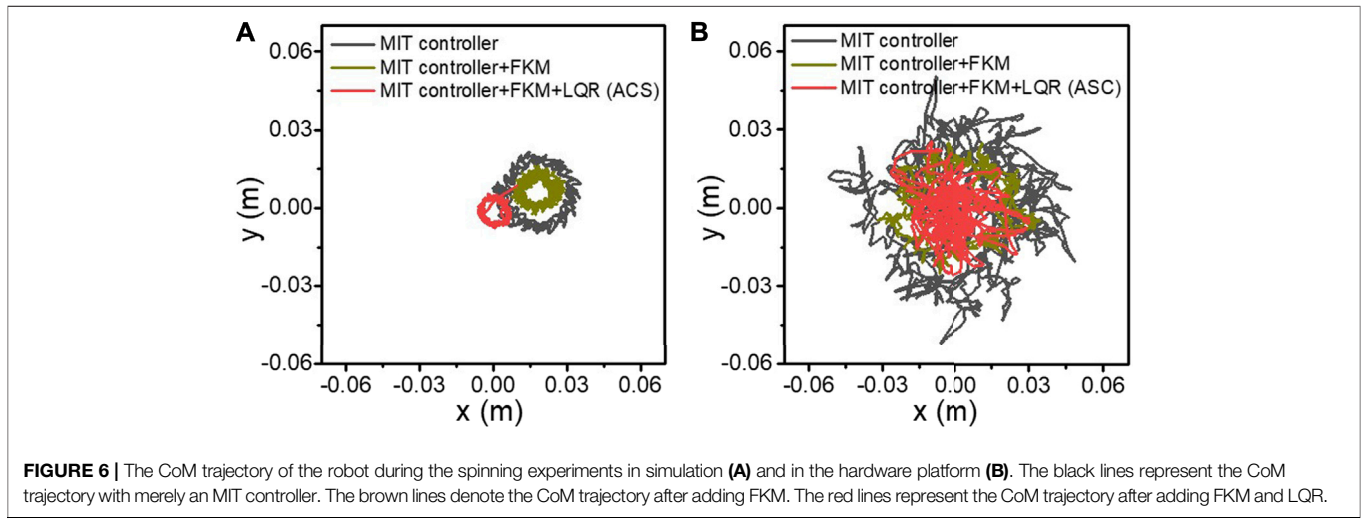

FIGURE 6 | The CoM trajectory of the robot during the spinning experiments in simulation (A) and in the hardware platform (B). The black lines represent the CoM trajectory with merely an MIT controller. The brown lines denote the CoM trajectory after adding FKM. The red lines represent the CoM trajectory after adding FKM and LQR.

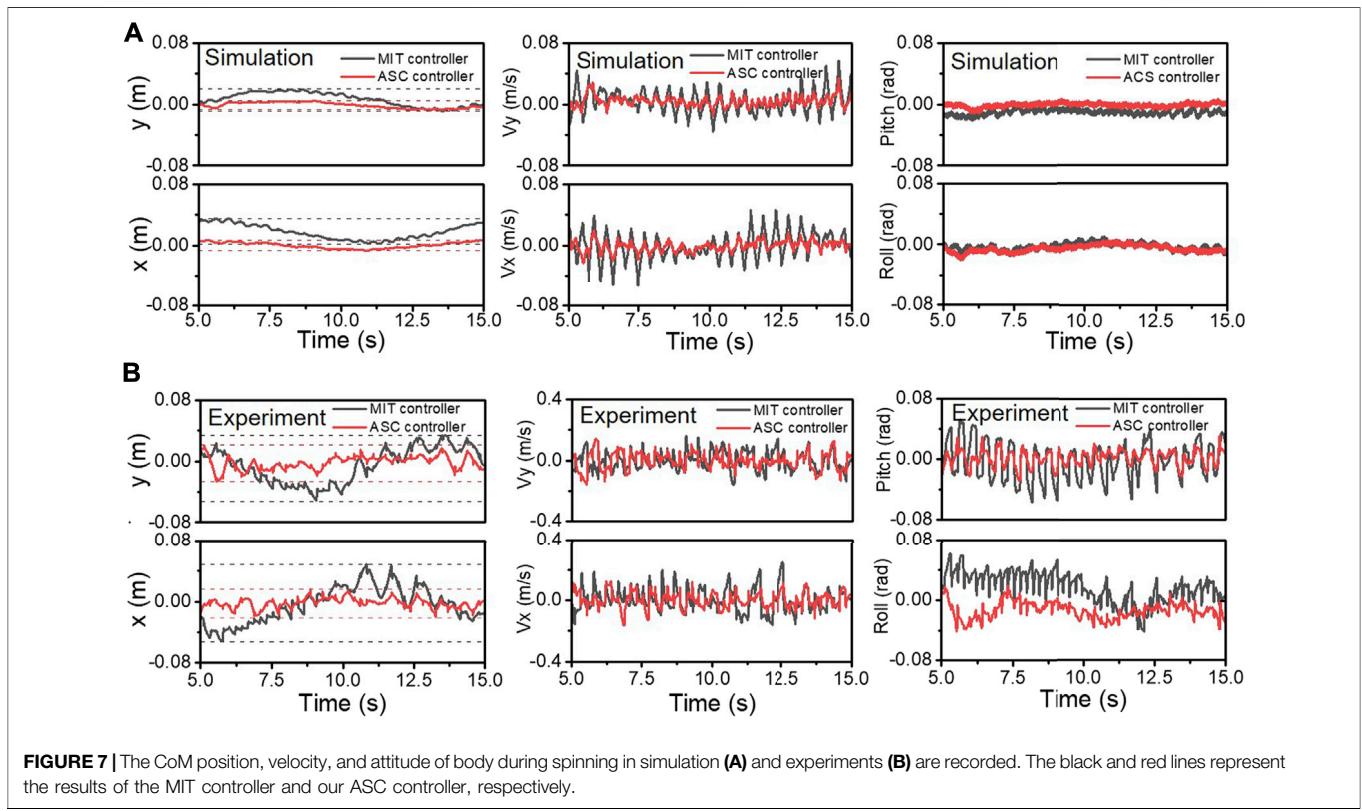

FIGURE 7 | The CoM position, velocity, and attitude of body during spinning in simulation (A) and experiments (B) are recorded. The black and red lines represent the results of the MIT controller and our ASC controller, respectively.

**Figure 8** shows a linear and angular acceleration phase diagram to demonstrate the stability improvement during spinning. The smaller the acceleration values in the $x$ and $y$ directions, the more stable the robot body. In simulation (**Figure 8A**), our ASC method reduces the variance from ($x$: $1.51 \times 10^{-1}$ (m/s$^2$)$^2$, $y$: $1.42 \times 10^{-1}$ (m/s$^2$)$^2$) to ($x$: $8.48 \times 10^{-2}$ (m/s$^2$)$^2$, $y$: $8.69 \times 10^{-2}$ (m/s$^2$)$^2$) for linear acceleration, and (Roll: $3.3 \times 10^{-3}$ (rad/s$^2$)$^2$, Pitch: $8.7 \times 10^{-2}$ (rad/s$^2$)$^2$) to (Roll: $1.1 \times 10^{-3}$ (rad/s$^2$)$^2$, Pitch: $1.3 \times 10^{-3}$ (rad/s$^2$)$^2$) for angular acceleration. In the experimentation (**Figure 8B**), the differences are not so obvious as in simulation, showing the variance from 0.933 (m/s$^2$)$^2$ to 0.784 (m/s$^2$)$^2$ for a linear acceleration of the $x$ direction, and (Roll: 0.142 (rad/s$^2$)$^2$, Pitch: 0.146 (rad/s$^2$)$^2$) to (Roll: 0.088 (rad/s$^2$)$^2$, Pitch: 0.084 (rad/s$^2$)$^2$) for angular acceleration, respectively. It is concluded that our work bound the acceleration during the spinning of the quadruped robot, showing better stability and smaller trajectory tracking errors.

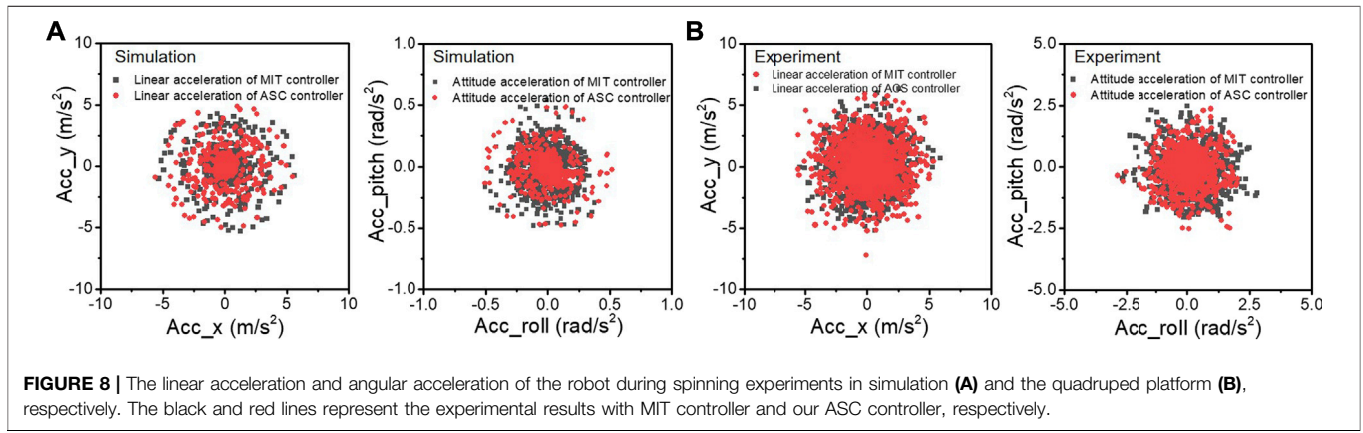

**FIGURE 8 |** The linear acceleration and angular acceleration of the robot during spinning experiments in simulation **(A)** and the quadruped platform **(B)**, respectively. The black and red lines represent the experimental results with MIT controller and our ASC controller, respectively.

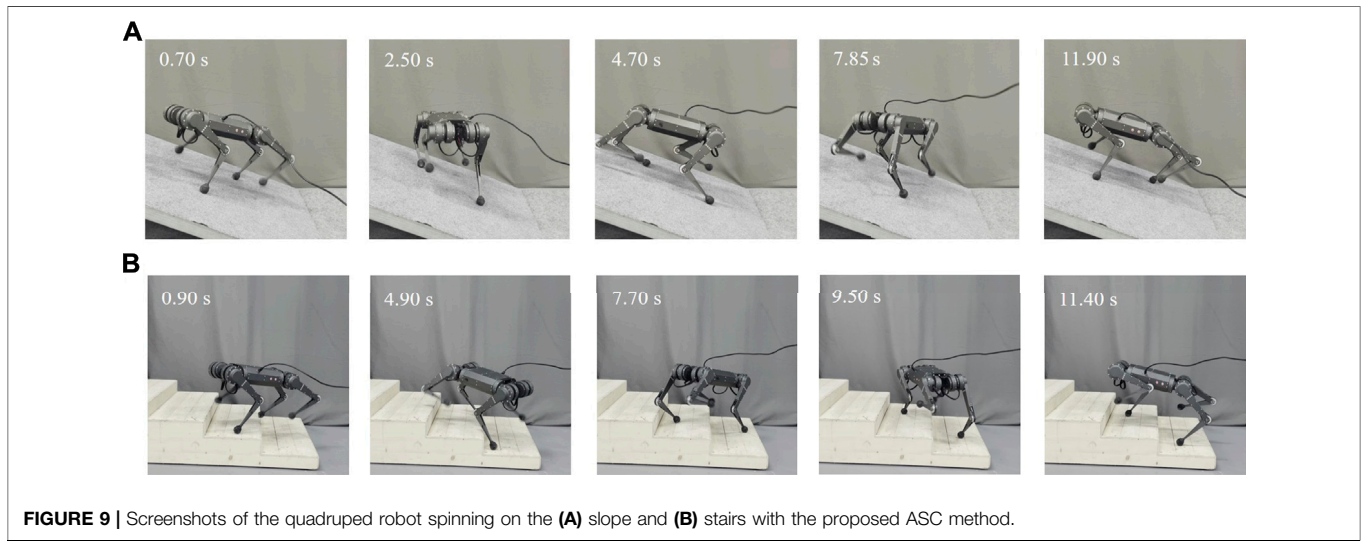

**FIGURE 9 |** Screenshots of the quadruped robot spinning on the **(A)** slope and **(B)** stairs with the proposed ASC method.

## 4.3 Experimental Validation of Spinning on Uneven Terrains

The spinning experiment is also conducted on slope and stair terrains to demonstrate the robustness of the proposed method. These terrains are also common scenes in human daily life. Compared with the flat ground spinning, these terrains bring gravity effect and obstacles as disturbance during spinning. By using the terrain estimation method mentioned above, our ASC method also showed robust performance on these terrains, as shown in **Figure 9** and **Supplementary Video S2**. As shown in **Figure 10**, the CoM trajectory and attitude of the robot body are recorded while spinning on the slope and stairs. A constant 0.7 *rad/s* spinning speed was maintained. With the terrain adaptation, the pitch angles changed periodically, ensuring the body is parallel to the slope. The small peaks are caused by the repeated steps. With our ASC method, the roll angle of the robot spinning on the slope has a small range from 0.352 to 0.165 rad, fluctuating around 0. The variance decreased from $5.8 \times 10^{-3}$ rad$^2$ to $9.8 \times 10^{-4}$ rad$^2$. For stairs, the performance is worse than that of the slope due to the discrete available footsteps and slipping and stumbling that occurs occasionally. With the ASC controller, the roll angle of the robot spinning on the stairs has a small range from 0.2597 to 0.2057 rad, and the variance decreases from $3.28 \times 10^{-3}$ rad$^2$ to $1.43 \times 10^{-3}$ rad$^2$. **Figures 10E,F** record the errors of position and angle of spinning on different terrains with varied spinning velocities of 0.8, 1.0, and 1.2 rad/s. The data are statistical results of 5 trials. In each trail, the robot spins at least 10 cycles corresponding to over 120 steps. The errors increase with larger angular velocities and the ground has the minimum error as expected. Other detailed velocity and acceleration data are in the Supplementary Materials. Overall, the effectiveness of the proposed method is demonstrated for improving both the accuracy and stability for spinning on slope and stairs.

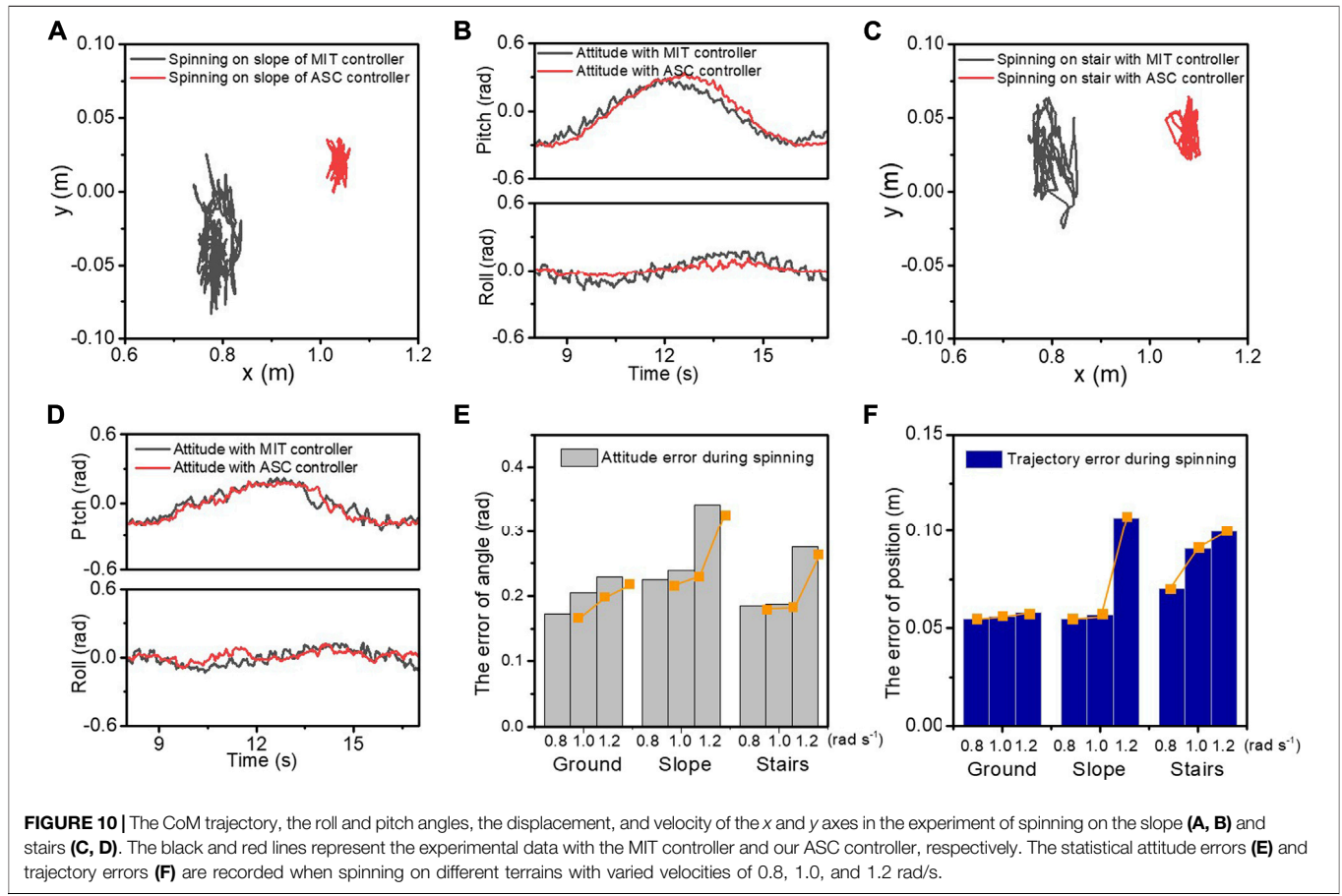

FIGURE 10 | The CoM trajectory, the roll and pitch angles, the displacement, and velocity of the x and y axes in the experiment of spinning on the slope (A, B) and stairs (C, D). The black and red lines represent the experimental data with the MIT controller and our ASC controller, respectively. The statistical attitude errors (E) and trajectory errors (F) are recorded when spinning on different terrains with varied velocities of 0.8, 1.0, and 1.2 rad/s.

## 5 CONCLUSION AND FUTURE WORK

The work presented in this study proposes an approach for terrain-perception-free but accurate spinning locomotion of a quadruped robot including a gait planner with spherical foot end effector modification, a CoM trajectory planner, and a LQR feedback controller. The roles of these three components are different and indispensable to accomplish the accurate spinning task. Specifically, the CoM trajectory planner is a modification of the traditional linear interpolation method. However, using only the linear interpolation method cannot maintain spinning on ground, and the robot falls after several turns of spinning. The foot end effector modification of the point-foot model error shows an improvement for the position error elimination during spinning. Besides the foot end effector rolling, an LQR feedback controller is added to further reduce the system errors. Experimental results on versatile terrains including flat ground, slope, and stairs are demonstrated. The radius of CoM trajectory and the variance of body state was reduced from 7.67 to 3.84 cm for ground through the comparison experimentation. Spinning is a type of agile locomotion and an indispensable part of turning. In fact, spinning can be treated as a special case of turning gait with a zero turning radius. According to our results, spinning can enlarge the defects of the model errors (foot end effector rolling in this work) or controllers. Thus, spinning can be treated as a standard evaluation method for testing the motion ability of legged robots, as proposed in the analysis of this study. Perception and path planning will be integrated into our framework in the future. By grasping a better understanding of the environment including the terrains and obstacle, accurate spinning ability has great potential to provide the legged robot with better adaptivity in narrow spaces.

## DATA AVAILABILITY STATEMENT

The original contributions presented in the study are included in the article/**Supplementary Material**; further inquiries can be directed to the corresponding authors.

## AUTHOR CONTRIBUTIONS

JL, HZ, DW, and YZ conceived the idea and designed the experiments. DW and HZ carried out the experiments and collected the data. HZ, JL, and YZ provided theory support.

JL, HZ, NB, YZ, AZ, LR, and DW discussed the results. HZ and DW wrote the manuscript, and JL, LR, NB, ZZ, and YZ contributed to editing the manuscript.

## FUNDING

This work was supported in part by the National Natural Science Foundation of China under Grant 51905251, in part by the National Key R&D Program of China (2019YFB1310402), and in part by AIRS project under Grant AC01202101023.

## SUPPLEMENTARY MATERIAL

The Supplementary Material for this article can be found online at: https://www.frontiersin.org/articles/10.3389/frobt.2021.724138/full#supplementary-material

**Conflict of Interest:** The authors declare that the research was conducted in the absence of any commercial or financial relationships that could be construed as a potential conflict of interest.

**Publisher's Note:** All claims expressed in this article are solely those of the authors and do not necessarily represent those of their affiliated organizations, or those of the publisher, the editors and the reviewers. Any product that may be evaluated in this article, or claim that may be made by its manufacturer, is not guaranteed or endorsed by the publisher.